\documentclass[10pt,twocolumn]{article}
\usepackage{iccv}
\usepackage{times}
\usepackage{epsfig}
\usepackage{graphicx}
\usepackage{amsmath}
\usepackage{amssymb}

\usepackage[pagebackref=true,breaklinks=true,letterpaper=true,colorlinks,bookmarks=false]{hyperref}
\usepackage{multirow}
\usepackage{bm}
\usepackage{color}
\usepackage[flushleft]{threeparttable}
\usepackage{enumitem}

\newcommand{\fig}{{Fig.}\@\xspace}
\newcommand{\tab}{{Tab.}\@\xspace}

\def\Vec#1{{\bm{#1}}}

 \iccvfinalcopy % *** Uncomment this line for the final submission

\def\iccvPaperID{2787} % *** Enter the ICCV Paper ID here

\ificcvfinal\fi
\begin{document}

%%%%%%%%% TITLE
\title{Fast Video Object Segmentation via Mask Transfer Network}

\author{Tao Zhuo$^\dag$, Zhiyong Cheng$^\ddag$, Mohan Kankanhalli$^\dag$ \\
$^\dag$National University of Singapore, $^\ddag$Shandong AI Institute \\
\tt \small zhuotao@nus.edu.sg, jason.zy.cheng@gmail.com, mohan@comp.nus.edu.sg}

%\author{Tao Zhuo\\
%National University of Singapore\\
%{\tt\small zhuotao@nus.edu.sg}
%% For a paper whose authors are all at the same institution,
%% omit the following lines up until the closing ``}''.
%% Additional authors and addresses can be added with ``\and'',
%% just like the second author.
%% To save space, use either the email address or home page, not both
%\and
%Zhiyong Cheng\\
%Shandong AI Institute\\
%{\tt\small jason.zy.cheng@gmail.com}
%\and
%Mohan Kankanhalli\\
%National University of Singapore\\
%{\tt\small mohan@comp.nus.edu.sg}}

\maketitle

\begin{abstract}
Accuracy and processing speed are two important factors that affect the use of video object segmentation (VOS) in real applications. With the advanced techniques of deep neural networks, the accuracy has been significantly improved, however, the speed is still far below the real-time needs because of the complicated network design, such as the requirement of the first frame fine-tuning step. To overcome this limitation, we propose a novel mask transfer network (MTN), which can greatly boost the processing speed of VOS and also achieve a reasonable accuracy. The basic idea of MTN is to transfer the reference mask to the target frame via an efficient global pixel matching strategy. The global pixel matching between the reference frame and the target frame is to ensure good matching results. To enhance the matching speed, we perform the matching on a downsampled feature map with 1/32 of the original frame size. At the same time, to preserve the detailed mask information in such a small feature map, a mask network is designed to encode the annotated mask information  with 512 channels. Finally, an efficient feature warping method is used to transfer the encoded reference mask to the target frame. Based on this design, our method avoids the fine-tuning step on the first frame and does not rely on the temporal cues and particular object categories.  Therefore, it runs very fast and can be conveniently trained only with images, as well as being robust to unseen objects.  Experiments on the DAVIS datasets demonstrate that MTN can achieve a speed of 37 fps, and also shows a competitive accuracy in comparison to the state-of-the-art methods.

\end{abstract}
\section{Introduction}

\begin{figure}[!t]
	\centering
	\includegraphics[width=1.0\columnwidth]{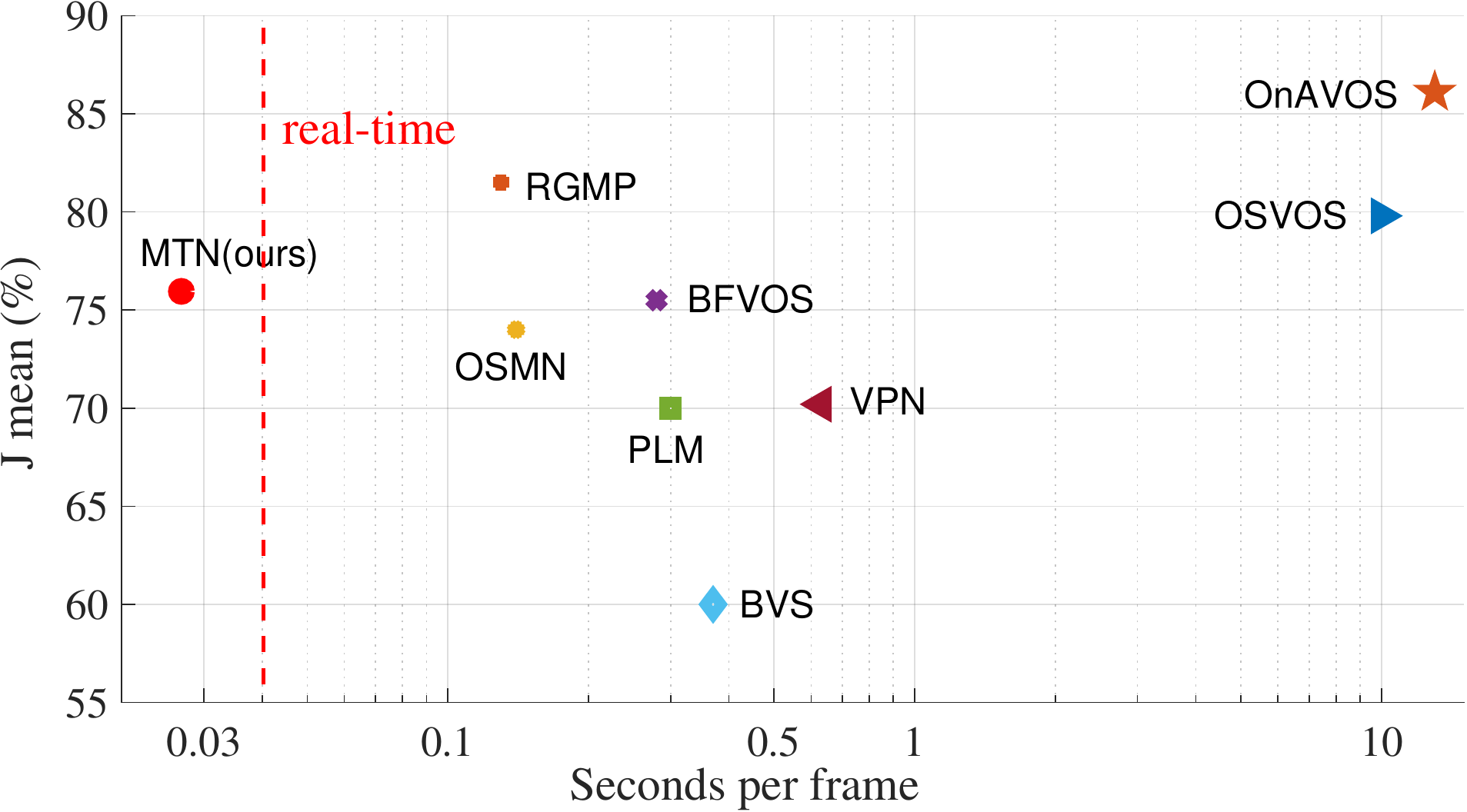}
	\caption{Accuracy (with Jaccard index metric $\mathcal{J}$) versus processing time on the DAVIS-16 validation set. Our method MTN achieves 37 fps (frames per second), which exceeds the real-time speed of 25 fps and is significantly faster than existing methods by a large margin. MTN also achieves comparable accuracy when compared to the state-of-the-art methods.}
	\vspace{-2ex}
	\label{fig_sa}
\end{figure}

Video Object Segmentation (VOS) is a fundamental problem that can be applied in many computer vision tasks including video stabilization, retrieval, summarization, editing and scene understanding.  In this paper, we focus on the semi-supervised VOS setting, which aims to segment a specific object in videos based on the annotated mask for this object in the first frame~\cite{CVPR2016_Perazzi}. In recent years, due to the advance of deep neural networks, great progress has been achieved, especially on the segmentation accuracy (\eg, OnAVOS~\cite{BMVC2017_Voigtlaender} achieves an accuracy of 86.1\% on the $\mathcal{J}$ mean metric~\cite{CVPR2016_Perazzi}). Although the processing speed has also been greatly improved (from 0.1 to 8 fps), it is still far from being real-time, which significantly limits their applications in practice. \fig \ref{fig_sa} shows the overall performance of the state-of-the-art semi-supervised VOS methods in terms of $\mathcal{J}$ mean and processing speed. As we can see, the segmentation accuracy of current methods is reasonable while the processing speed still has a big gap to reach real-time.

%Many existing semi-supervised algorithms~\cite{BMVC2017_Voigtlaender,CVPR2017_Caelles,CVPR2017_Perazzi,ICCV2017_Cheng,arxiv2017_Khoreva,CVPR2018_Yang} need to further fine-tune the pre-trained model on the first frame to achieve a high accuracy, which is often time-consuming and not practical in real applications. For example, the recent method OSVOS~\cite{CVPR2017_Caelles} requires more than 10 minutes for the first frame fine-tuning step on each video.

%Given the first frame with an annotated object mask, the goal of semi-supervised method is to automatically segment the specific objects in videos.
The recent semi-supervised VOS methods mainly depend on two strategies: \emph{propagation-based} and \emph{detection-based}. The former approaches formulate the VOS task as a mask propagation problem from the first frame to the subsequent frames~\cite{CVPR2017_Perazzi,arxiv2017_Khoreva}. 
To leverage the temporal consistency between two adjacent frames, many \emph{propagation-based} methods~~\cite{CVPR2017_Perazzi,arxiv2017_Khoreva,ICCV2017_Cheng} often transfer the mask of the previous frame to the current frame with optical flow estimation. However, because the computational cost of dense and accurate optical flow estimation is often expensive, their processing speed is severely limited. For example, SFL \cite{ICCV2017_Cheng} simultaneously predicts pixel-wise object segmentation and optical flow in videos, it runs at 7.9 seconds per frame.

The \emph{detection-based} strategy \cite{CVPR2017_Caelles,ICCV2017_Yoon,CVPR2018_Chen,ECCV2018_Hu} addresses the VOS task as a pixel-level object retrieval problem with the annotation of the object mask in the first frame. Those methods often adopt a two-stage learning strategy that first trains an offline model to extract the semantic regions on generic objects, and then makes the offline model focus on the particular object by fine-tuning it on the first frame with annotated object mask. The problem is that the first frame fine-tuning step is often time-consuming. A typical example is the recent method OSVOS~\cite{arxiv2017_Khoreva} that can run at 10 fps, but it needs more than 10 minutes for the first frame fine-tuning step on each video to achieve a high accuracy. 

To speed up the video object segmentation, researchers recently have paid more attention to improve the VOS speed while sacrifice some accuracy, such as OSMN~\cite{CVPR2018_Yang}, BFVOS~\cite{CVPR2018_Chen}, VM~\cite{ECCV2018_Hu} and RGMP~\cite{CVPR2018_Oh}. Although great progress has been achieved so far (\eg, RGMP can run at approximately 8 fps), they are still far from being real-time. In this paper, we aim to develop a real-time yet still accurate VOS method. To achieve this goal, we propose a novel method called Mask Transfer Network (MTN), which segments the target object by transferring the annotated mask from the first frame to the target frame with an  efficient global pixel matching strategy. Different from the existing approaches, our method enjoys two main merits: \emph{avoids the time-consuming first frame fine-tuning step} and \emph{does not rely on particular object categories and temporal cues.} Therefore, it is robust to unseen objects and can be trained on the general image object segmentation datasets. 

%In this paper, we target at developing a real-time yet still accurate VOS method. To achieve this goal, we propose a novel semi-supervised VOS method called Mask Transformation Network (MTN), which can automatically avoid the time-consuming first frame fine-tuning step. Our proposed approach MTN can achieve a very fast speed (37 fps) with a competitive accuracy (75.3\% on the DAVIS-16 trainval set, 50 videos). 

%In this paper, to develop a real-time yet still accurate VOS method without the time-consuming first frame fine-tuning, we propose to segment the target object by transferring the annotated mask from the first frame to target frame with an efficient global matching method. Different from most of the existing \emph{detection-based} approaches~\cite{CVPR2017_Caelles,ICCV2017_Yoon,CVPR2018_Chen,ECCV2018_Hu}, the pre-trained model of our approach does not rely on the particular object. Therefore, the first frame fine-tuning step could be automatically omitted in our method. 

Specifically, given a video sequence with an annotated object mask on the first frame (\ie, reference frame), the key idea of our method is to transfer the annotated object mask to the target frame by an efficient global pixel matching strategy between the reference frame and the target frame. For an arbitrary video frame, the target object may be located at any position. Consequently, the corresponding locations of an annotated object should be globally matched over the target frame. However, performing the global pixel matching step on all pixels of the original image size will be very time-consuming. 
To solve this problem, we propose to efficiently match pixels on a feature map of a much smaller size with high dimensions.

\begin{figure*}[!t]
	\centering
	\includegraphics[width=0.9\textwidth]{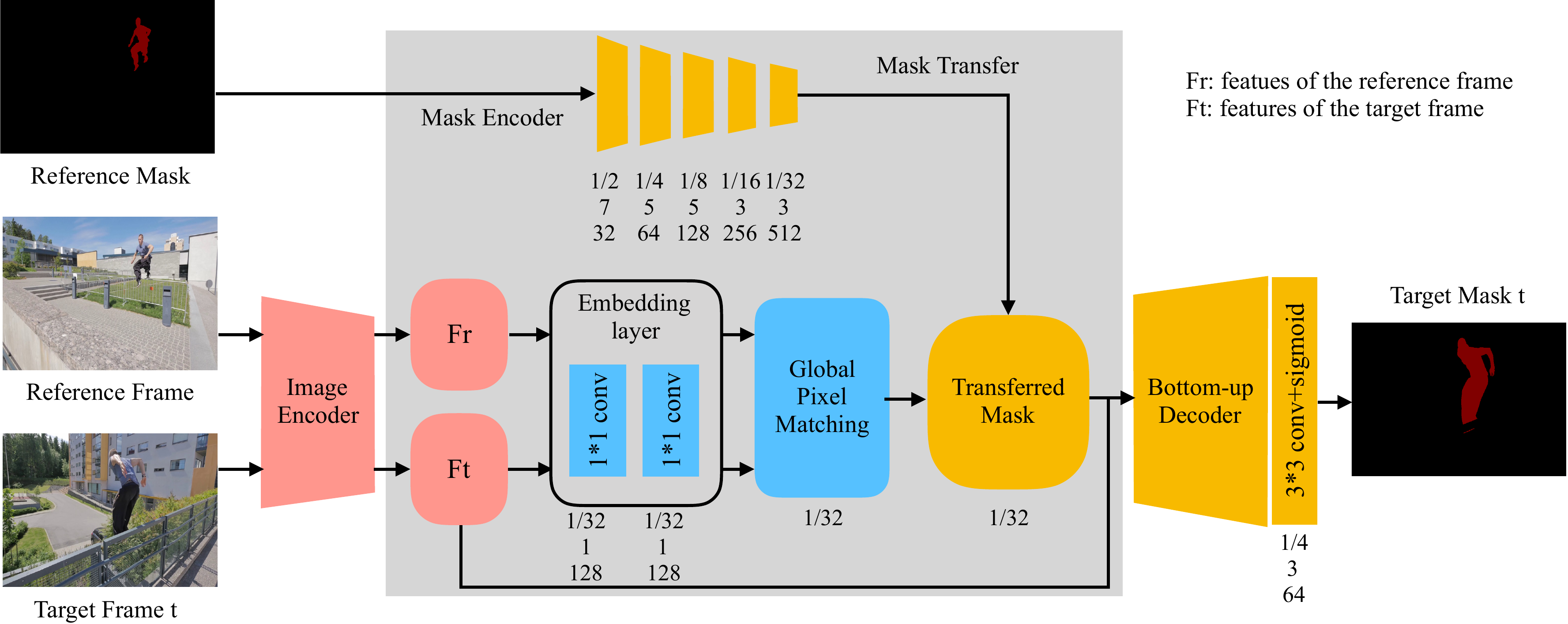}
	\caption{The architecture of the proposed method MTN. The relative spatial sizes and channel dimensions of feature maps are denoted below each module.}
	\label{fig_framework}
\end{figure*}

As shown in Fig.~\ref{fig_framework}, to speed up the global pixel matching step, we encode the image features of the reference frame and target frame into a very small size (\eg, 1/32 size of the input image). Meanwhile, in order to transfer the reference mask to the target frame on the downsampled feature map (\ie, 1/32) and simultaneously preserve detailed mask information (\eg, object boundaries), we build a mask encoder network (5 convolutional layers with a stride of 2) to encode the reference mask. To this end, the encoded mask of the reference frame can be effectively transferred to the target frame by using a simple warping operation based on the pixel matching results. Finally, the concatenated features of the encoded image features of the target frame and the transferred mask are fed into the last layer of a bottom-up decoder for the target object segmentation. 

Empirical studies on the DAVIS benchmark dataset~\cite{CVPR2016_Perazzi} show that the proposed method MTN can achieve a speed of 37 fps on images of $854 \times 480$ pixels, which is much faster than all the existing methods. Even when compared to  RGMP~\cite{CVPR2018_Oh} - one of the top fastest methods (less than 8 fps) - on the same platform, MTN is still $4.8\times$ faster. At the same time, MTN can also achieve competitive accuracy from different perspectives: (1) The proposed method MTN does not rely on temporal cues, and thus it can be trained on general image object segmentation datasets without any annotated video sequences. By training the proposed network on two image object segmentation datasets (\ie PASCAL VOC~\cite{IJCV2010_Everingham} and MSRA10K~\cite{TPAMI2015_Cheng}), MTN achieves 75.3\% of $\mathcal{J}$ on the DAVIS-16 trainval set (50 videos), which is highly competitive to the state-of-the-art methods (see Sec. \ref{subsec_exp_train}). (2) As MTN does not rely on particular object categories, it can be used to segment unseen objects. Compared to other recently developed mask transfer methods, MTN significantly improves the accuracy by a large margin of $14.9\%$ (see Sec. \ref{subsec_unseen}). (3) Compared to the state-of-the-art methods on the DAVIS-16 validation set and multi-object segmentation on the DAVIS-17 validation set, MTN achieves competitive accuracy (see Sec. \ref{subsec_val} and \ref{subsec_multi}).

\section{Related Work}

{\bf Unsupervised methods.} 
Unsupervised VOS methods \cite{BMVC2014_Faktor, CVPR2015_Fragkiadaki,CVPR2017_Tokmakov,CVPR2017_Jain,CVPR2017_Koh,CVPR2018_LiSiyang} aim to automatically segment prominent objects without any user annotations.
These methods usually rely on the visual saliency cues such as motion and long-term trajectories \cite{BMVC2014_Faktor, CVPR2015_Fragkiadaki}.
Based on motion cues, recent methods \cite{CVPR2017_Tokmakov,CVPR2017_Jain} often detect the moving regions that indicate semantic objects  with deep learning networks, by jointly using optical flow and object proposal methods.
On the other hand,
long-term trajectory-based methods \cite{BMVC2014_Faktor,CVPR2015_Fragkiadaki,CVPR2017_Koh} depend on the temporal consistency of pixels, superpixles or object proposals, with the assumption that pixels with consistent trajectories are foreground objects and the rest pixels are background.
Due to the lack of information about the target object, however, unsupervised methods often fail to accurately segment a specific object in videos. 
Besides, they also easily suffer from the motion confusions between the dynamic background and other objects \cite{CVPR2017_Tokmakov,CVPR2017_Jain}, resulting in poor performance. Therefore, we mainly focus on the semi-supervised approach in this paper.

{\bf Semi-supervised methods.} 
Given the first video frame with annotated object masks, semi-supervised methods \cite{CVPR2018_Li,ICCV2017_Yoon,CVPR2017_Caelles,CVPR2017_Perazzi,CVPR2018_Yang,CVPR2018_Chen,arxiv2017_Khoreva} aim to segment specific objects across the entire video sequence. To deal with fast motion and heavy occlusion, many approaches~ \cite{CVPR2017_Caelles,CVPR2017_Perazzi,BMVC2017_Voigtlaender,arxiv2017_Khoreva} first train an offline model to generate generic object proposals, and then fine-tune the offline model on the first video frame for particular target object segmentation. Although this fine-tuning strategy significantly improve the accuracy, its expensive computational cost makes those algorithms unsuitable for practical applications. 

Recently, many approaches~ \cite{CVPR2018_Chen,CVPR2018_Oh,ICCV2017_Yoon} are developed for fast VOS by avoiding the first frame fine-tuning step. FAVOS \cite{CVPR2018_Cheng} adopts a part-based tracking method to predict bounding boxes of object parts and then segment the target object with a segmentation network. OSMN~\cite{CVPR2018_Yang} uses a network modulation approach to manipulate intermediate layers of the segmentation network. RGMP~\cite{CVPR2018_Oh} proposes a hybrid model that fuses the mask detection and propagation in a Siamese encoder-decoder network. 
%By training the network on multiple datasets and synthetic videos with backpropagation through time, 
RGMP achieves good segmentation accuracy and it can run at approximately 8 fps. However, due to the inefficient network architecture design, the processing speed of existing algorithms is still far from being real-time. In this work, we proposed to apply a global pixel matching strategy for fast and accurate VOS. More details on several matching-based approaches are discussed in Sec. \ref{subsec_discuss}.

\section{Mask Transfer Network}

\begin{figure}[!t]
	\centering
	\includegraphics[width=1.0\columnwidth]{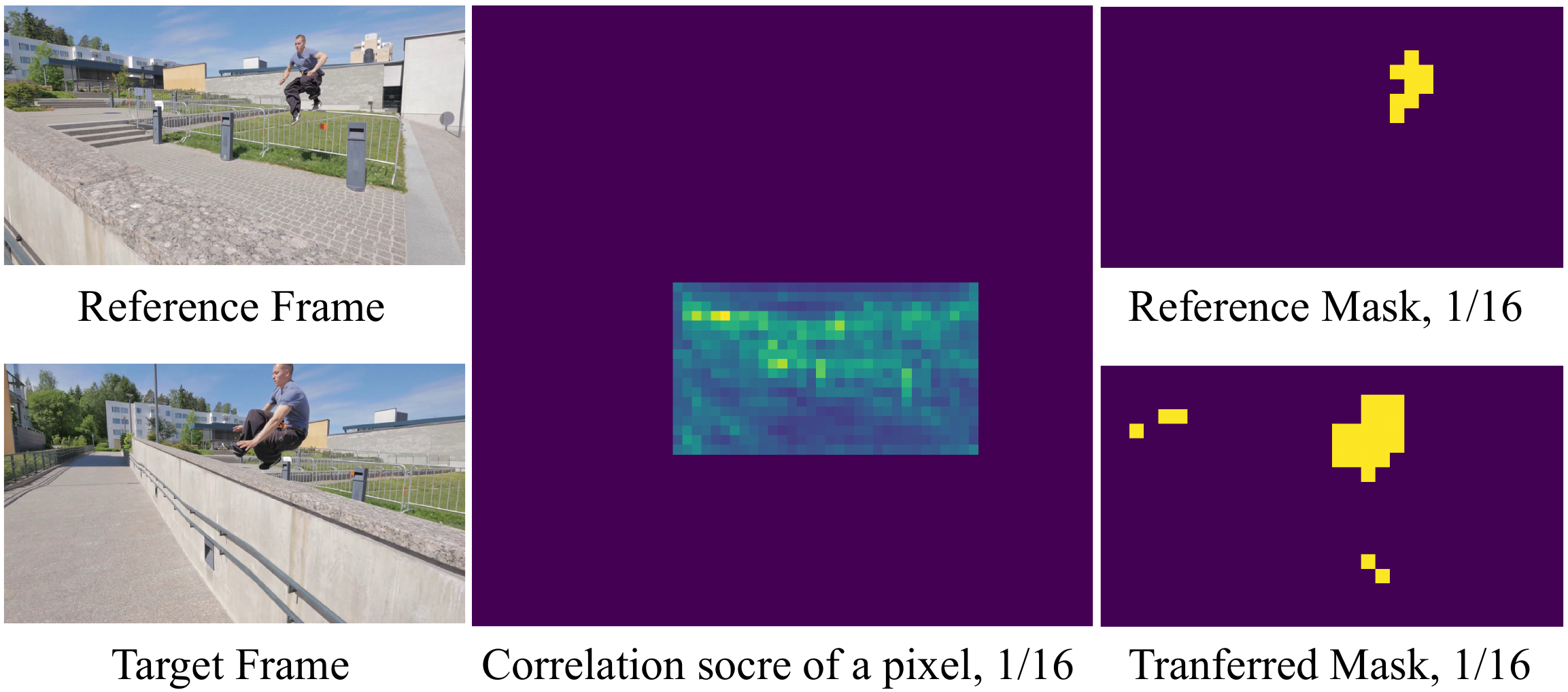}
	\caption{An example of the mask transfer. In order to clearly illustrate the mask transfer module, we demonstrate the transferred mask on the 1/16 sized image with one channel (reference mask is directly resized to 1/16 scale), which can provide a coarse location of the target object. Besides, the correlation score of a pixel with $d_m^2$ dimensions is resized to a $d_m \times d_m$ score map, in which the pixel with maximum correlation score value indicates its corresponding pixel in the target frame.}
	\label{fig_trans}
\end{figure}

Given a reference frame (\ie the first frame of a video) with an annotated object mask, our goal is to achieve fast and accurate object segmentation over the entire video sequence. The key idea of our method is to transfer the annotated object mask to the target frame based on an efficient global pixel matching strategy between the reference frame and the target frame. However, performing global pixel matching on the original size of frames will be very time-consuming. Thus we need to downsample the image into a smaller size for fast processing. At the same time, the annotated mask needs to be downsampled accordingly for mask transfer. A dilemma here is that the reference mask cannot be directly resized to a very small size, which will cause significant information loss of the reference mask (\eg object boundaries, see \fig \ref{fig_trans}). Moreover, for objects which are already of very small sizes, this strategy is inapplicable either. To solve this problem, we apply a mask network to preserve detailed mask information for accurate mask transfer.

\subsection{Network Architecture}
The architecture of the proposed method MTN is shown in \fig \ref{fig_framework}. 
The key modules of our method include an \emph{image encoder} for feature extraction, a \emph{global pixel matching} module and a \emph{mask encoder} for reference mask transfer, and a \emph{bottom-up decoder} for the target object segmentation. In the next, we will introduce those modules in sequence.

{\bf Image encoder.}
The image encoder is used to extract the features of input RGB images. To match the extracted image features of the reference frame and the target frame in the same feature space, a Siamese network~\cite{ICML2015_Koch} with shared weights is used as the image encoder. 
In our implementation, the ResNet50~\cite{CVPR2016_He} is used as the image encoder and the weights are initialized from the pre-trained model on ImageNet~\cite{CVPR2009_Deng}. Notice that the initialized parameters can be fine-tuned during the training of MTN for better performance. For simplicity, we fixed those parameters in our experiments.

{\bf Pixel matching.}
Mask transfer is built upon the pixel correspondences between the reference frame and the target frame, which can be achieved by a global pixel matching method and a simple feature warping operation~\cite{CVPR2018_Sun}. Our pixel matching algorithm is inspired by the optical flow estimation methods~\cite{ICCV2015_Dosovitskiy,CVPR2018_Sun}. In those methods, the pixel displacement is usually small, since optical flow estimation is operated on two adjacent frames. However, in VOS, the pixel displacements could be very large because the target object could be located at any position in the target frame. Consequently, the pixel matching needs to be operated globally to cover the whole feature map in VOS. To speed up the global pixel matching step, we downscale the feature map into 1/32 of the original size, reducing the number of pixels for matching to $(1/32)^2$ of the original magnitude. Hence, the processing speed can be greatly accelerated. Besides, we adopt an embedding layer for the purpose of robust pixel matching. It is expected that pixels are much easier to be matched in a learned embedding space~\cite{CVPR2015_Schroff}. As illustrated in \fig \ref{fig_framework}, our embedding layer consists of two convolutional layers with 128 output channels. Notice that the processing speed of matching step is further improved by compacting the image feature dimension from 2048 to 128.

{\bf 1) Local pixel matching.}
In deep learning based optical flow estimation~\cite{ICCV2015_Dosovitskiy,CVPR2018_Sun}, efficient pixel matching can be achieved with a correlation layer \cite{ICCV2015_Dosovitskiy} that measures the patch similarity between two images. Formally, let $x^r_i$ denote the $i$-th pixel in the reference frame $I_r$, $x^t_j$ denote the $j$-th pixel in the $t$-th target frame $I_t$. Given the learned embedding features $f(x^r_i,d)$ and $f(x^t_j,d)$ of two patches (with a patch size $d$) centered at $x^r_i$ in $I_r$ and $x^t_j$ in $I_t$, 
the ``correlation score'' $c(x^r_i, x^t_j)$~\cite{ICCV2015_Dosovitskiy} of two pixels $x^r_i$ and $x^t_j$ by the patch similarity is computed by the cross-correlation as:
\begin{equation}
\centering
c(x^r_i, x^t_j) = \sum_{\Vec{o} \in [-d,d] \times [-d,d]}{\langle f(x^r_i,\Vec{o}), f(x^t_j,\Vec{o}) \rangle}
\label{eqn_corr}
\end{equation} 
where $\Vec{o}$ is the pixel displacement and patch size is $D=2d+1$. Since the optical flow estimation methods~\cite{ICCV2015_Dosovitskiy,CVPR2018_Sun} often assume that only small pixel displacements exist between two adjacent images, the patch size $d$ is often set to a small value for local pixel matching. 
Note that Eq. \ref{eqn_corr} computes the patch similarity of two input features, and thus it does not involve any trainable parameters.

{\bf 2) Global pixel matching.}
As the target object can be located at any position in target frames. Therefore, we design a global pixel matching strategy to compute the similarity of all pixels between the reference frame and the target frame. 

Let $w$ and $h$ denote the width and height of a feature map, respectively. For global pixel matching, the maximum displacement $d_m$ of a pixel is set to $d_m = max(w, h)$, then the corresponding patch size is $D=2d_m+1$. The output of the correlation layer denotes all of the patch similarity scores for a given pixel. The size of the output is $w \times h \times D^2$. \fig \ref{fig_trans} illustrates the resized $D \times D$ correlation score map of a pixel on the 1/16 sized image.

The global pixel matching step is to find the most similar pixels with the maximum correlation score over the whole target image. Let $x_i^r$ be the $i$-th pixel in the reference frame $I^r$, $x_j^t$ denote the $j$-th pixel in the target frame $I^t$, and then $S_i^t=\{c(x^r_i, x^t_1), c(x^r_i, x^t_2),\cdots, c(x^r_i, x^t_N)\}$ be the correlation scores between $x_i^r$ and the image features of all the pixels in $I^t$, where $N$ is the total number of pixels on the feature map. Let $h$ be the index of the pixel with maximum correlation score in $S_i^t$, the displacement $(dx, dy)$ between a pixel $x_i^r$ and its corresponding pixel $x_j^t$ is computed as:
\begin{equation}
\centering
\begin{split}
    dx = mod(h, 2d_m+1) - d_m \\
	dy = div(h, 2d_m+1) - d_m
\end{split}
\label{eq_dis}
\end{equation}
where $dx$ and $dy$ represent the pixel displacement in horizontal and vertical direction, respectively. $mod$ denotes the mod operator and $div$ represents the division operator. From the description, it can bee seen that if we perform the global pixel matching on the original size of frames, it would be very time-consuming. Therefore, we downscaled the frame size into 1/32 of the original size as mentioned above. The problem is that the reference mask also needs to be downscaled accordingly for transfer, which will be described in the below.

{\bf Mask encoder and transfer.}
\label{subsec_transfer}
As pixel correspondences are computed on a smaller size of the input images, the reference mask also needs to be scaled to the same size for mask transfer. However, directly resizing the original mask to a very small size  will lose detailed information of the reference mask \cite{CVPR2018_Chen,ECCV2018_Hu} (such as image boundaries, see \fig \ref{fig_trans}). Especially for objects of very small sizes, this strategy is obviously inapplicable. To address the problem, we build a mask encoder to preserve the detailed information of the reference mask during the downscaling processing. As shown in Fig.~\ref{fig_framework}, the mask encoder consists of 5 convolutional layers with a stride of 2, which encodes the original mask into a feature map with 1/32 size of the input image and high dimensions (512 channels).
This mask encoder can effectively preserve the detailed mask information (according to the experimental results). Finally, based on the pixel matching results, a feature warping method~\cite{CVPR2018_Sun} (\ie assigning values to corresponding pixels based on computed pixel displacements) is used to transfer the encoded reference mask to the target frame. \fig \ref{fig_trans} illustrates the effectiveness of the mask transfer on a downsampled size with single channel representation. Although the appearances of the reference frame and the target frame are very different, the important location cues of the target object are still obtained by the proposed mask transfer method. Because of the proposed global pixel matching strategy and the mask encoder design, we can achieve accurate and fast target object mask estimation in the downsampled feature space. 
 
{\bf Bottom-up decoder.} 
After representing the target object mask at a downsampled scale, the next step is to implement the segmentation of the target object at the original size. This step is achieved by several de-convolutional layers (\ie 5 layers) in the bottom-up decoder module. Specifically, the encoded image features of the target frame and the transferred mask features are concatenated together, and then fed into the last layer of a global convolutional block \cite{CVPR2017_Peng} to extract the semantic features of the target object. Next, several residual-based boundary refinement blocks \cite{CVPR2017_Peng} are used to generate a score map with 64 channels. More details about the decoder processing are discussed in \cite{CVPR2017_Peng}. Based on an additional convolutional layer, a single channel score map is further produced as the foreground probability map. Since the bottom-up decoder is implemented on a downsampled image feature map ($1/4$ size of the input image), we upsample foreground probability map to the original image size by an interpolation operation.

\subsection{Model Training}
\label{subsec_train}
Given a reference frame (\ie the first frame) with an annotated object mask (\ie the reference mask which is manually labeled), MTN can automatically segment the target object in the subsequent frames. As shown in Fig. \ref{fig_framework}. all the target frames are matched with the reference frame independently, which means that MTN does not rely on temporal cues. Therefore, MTN can be trained from images without any annotated video sequences, see Sec. \ref{subsec_exp_train}.

During the training stage, a pair of \emph{images} or \emph{video frames} with corresponding masks can be used in MTN for training.  We adopt the dice coefficient loss \cite{IC3V_Milletari} in image object segmentation to measure the overlap between predicted mask and the ground-truth. Specifically, the employed dice loss is defined as:
\begin{equation}
loss = 1 - \frac{2 \sum_{i=1}^{N}{g_i * p_i} + \epsilon}{\sum_{i=1}^{N}{g_i^2}+\sum_{i=1}^{N}{p_i^2} + \epsilon}
\end{equation}
where $g_i$ and $p_i$ are the ground truth label and  predicted foreground probability of the pixel $x_i$, respectively. $\epsilon$ is a smooth factor and we set it to 1 in our implementation.
%In addition, we train the entire network end-to-end with the Adam optimizer \cite{ICLR2015_Kingma} and image patch size $512 \times 512$. 
%The learning rate is set to $10^{-5}$. %We train the network for $240,000$ iterations.

\subsection{Multi-object segmentation}
To deal with the multi-object segmentation problem, we first compute the foreground probability of each single object independently, and then we compute the background probability of the target frame as:
\begin{equation}
p_{i,0} = 1 - \frac{1}{M}{\sum_{m=1}^{M}p_{i,m}}
\end{equation}
where $p_{i,m}$ denotes the foreground probability of the object $m$ at the $i$-th pixel $x_i$, $m=0$ represents the background and $M$ indicates the number of objects. The pixel label with the maximum output probability is used as the final target object segmentation.

\subsection{Comparison with Matching-based Methods}
\label{subsec_discuss}
A big advantage of our method MTN is that it does not need any model fine-tuning for VOS, which can greatly save the processing time. Besides, the encoder-decoder network is meticulously designed to accelerate the processing speed and can also ensure the segmentation accuracy (see \tab~\ref{tbl_davis2016_all}). In the following, we discuss the main differences between our MTN with the approaches which also adopt the pixel-wise matching strategy.

PLM~\cite{ICCV2017_Yoon} combines the pixel-wise matching module with object proposals in a unified network. It uses a two-stage learning strategy that requires fine-tuning of the pre-trained model on the first video frame to segment the target object. Unlike PLM, our proposed approach MTN does not rely on the first frame fine-tuning. 
BFVOS~\cite{CVPR2018_Chen} and VM~\cite{ECCV2018_Hu} use a nearest neighbour classifier to match pixels between two video frames. To reduce the computation cost of the matching step, they use the image feature maps of a downsampled scale (\ie, 1/8 size of the input image), and then they directly resize the reference mask to the same scale for target mask prediction. In contrast, our method performs the pixel matching on a feature map with a smaller size, \ie, 1/32 size of the input image. As discussed in Sec. \ref{subsec_transfer}, without the mask encoder module, BFVOS and VM cannot perform the matching step on a very small size, because direct resizing of the reference mask will cause significant information loss, leading to large segmentation accuracy degradation. Besides, instead of using the pixel-wise distance measurement in BFVOS and VM, our approach MTN employs a correlation layer~\cite{ICCV2015_Dosovitskiy} to measure the patch similarity centered at a pixel, in which the spatial information of the feature map is considered, and thus more accurate matching results can be obtained.

%BFVOS~\cite{CVPR2018_Chen} formulates the semi-supervised VOS as a pixel-wise retrieval problem in an embedding space, which is learned by a nearest neighbour classifier and triplet loss. During inference, an online adaption component is used for model update. To reduce the computation cost, BFVOS matches pixels of two image feature maps at a small scale (\ie 1/8 times of the input image size) and it directly resizes reference mask to the same scale for target mask prediction. 
%
%However, without the 
% Different from BFVOS, MTN does not use the nearest neighbour classifier and online model update. 
%
%VM~\cite{ECCV2018_Hu} adopts a similar strategy to BFVOS that uses a nearest neighbour classifier to match pixels between the reference frame and the target frame in a learned embedding space. Unlike BFVOS, VM directly segment the target object from produced matching score without triplet loss. Instead of using the pixel-wise distance measurement in BFVOS and VM, the proposed MTN approach uses a correlation layer~\cite{ICCV2015_Dosovitskiy} to globally match pixels of encoded image features, in which the spatial information is considered. Moreover, the reference mask in MTN is encoded at a feature map with a smaller size (\ie, $1/32$ times of the input image size) and high dimension (\ie, 512 dimensions). This is quite different from BFVOS and VM that directly resize the reference mask for inference.

\section{Experiments}
In the following， we first introduce the implementation details, and then report and analyze the performance of our method from different perspectives. 

\subsection{Implementation Details}
We use the ResNet50 network (pre-trained on ImageNet) as the image encoder to extract image features. The parameters of the ResNet50 are fixed. During the training stage, the size of image patches is $512 \times 512$ and the learning rate is set to $10^{-5}$ with the Adam optimizer \cite{ICLR2015_Kingma}.
The bottom-up decoder is designed based on the semantic image segmentation module \cite{CVPR2017_Peng}. In our experiments, the proposed method is implemented and evaluated on a single NVIDIA GeForce 1080 Ti GPU.

\subsection{Results on the Entire DAVIS-16 Dataset}
\label{subsec_exp_train}
To demonstrate the advantage that our method can be trained on images without any annotated video sequences, we train the proposed network MTN on two image segmentation datasets, \ie PASCAL VOC~\cite{IJCV2010_Everingham} and MSRA10K~\cite{TPAMI2015_Cheng}. Then we evaluate the performance of trained MTN on the entire DAVIS-16 trainval set~\cite{CVPR2016_Perazzi}, in which all 50 videos (from both training and validation sets) are completely blind to PASCAL VOC and MSRA10K. PASCAL VOC is a popular semantic image segmentation dataset with instance-level object masks, which consists of 2913 images with 20 different object categories.
MSRA10K is a salient object segmentation dataset that consists of 10,000 images.
DAVIS-16 benchmark dataset focuses on the single object segmentation. It contains 50 high resolution ($854 \times 480$) videos with pixel-level binary mask annotations. This dataset has 30 videos for training and 20 for validation. %Typical challenging conditions in DAVIS-16 dataset include drastic appearance change, dynamic background, fast motion, motion blur and heavy occlusion.

In our experiment, we randomly choose an image with a single object mask from PASCAL VOC as a training sample. To simulate two video frames as the input to MTN for training, we augment the images with a set of random transformations (\ie, horizontal flipping, brightness, contrast, scaling, affine transformation, and center cropping) to generate a pair of images. During training, we use a fixed learning rate of $10^{-5}$, and the number of iterations is set to 400. To improve the diversity of training samples, we further fine-tune the trained model on the MSRA10K dataset with a fixed learning rate of $5 \times 10^{-6}$ and 3 iterations.

\begin{table}[!t]
	\centering
	\small
	\caption{Comparisons on the DAVIS-16 trainval set (50 videos). FT: first frame fine-tuning; PF: previous frame for mask propagation; OF: optical flow.}
	
	\resizebox{1\columnwidth}{!}{
		\begin{tabular}{l|ccc|c|c|c}   \hline
			& FT   & PF      & OF  & Time (s) & $\mathcal{J}$ mean &  $\mathcal{F}$ mean  \\ \hline \hline
			MSK \cite{CVPR2017_Perazzi}    & Y   & Y       & Y   &  12  & 80.3 & 75.8      \\
			\underline{\bf MTN (P+M)}      &     &         &     &  \underline{0.027}  & \underline{75.3} & \underline{76.1}   \\
			\underline{\bf MTN (P)}        &     &         &     &  \underline{ 0.027} & \underline{72.4} & \underline{74.9}    \\ 			
			VPN \cite{CVPR2017_Jampani}    &     & Y       & Y   & 0.63  & 75.0 & 72.4     \\  			
			CTN \cite{CVPR2017_Jang}       &     & Y       & Y   &  30  & 75.5 & 71.4      \\ 
			OFL \cite{CVPR2016_Tsai}       &     & Y       & Y   & 120 & 71.1 & 67.9    \\  
			BVS \cite{CVPR2016_Nicolas}    &     & Y       &     & 0.37  & 66.5 & 65.6     \\		
			JMP \cite{SIGGRAPH2015_Fan}    &     & Y       & Y   & -  & 60.7 & 58.6 \\
			FCP   \cite{ICCV2015_Perazzi}  &     & Y       & Y   & 14.5  & 63.1 & 54.6      \\  \hline 

		\end{tabular}
		}
	\label{tbl_davis2016_all}
    \vspace{-2ex}
\end{table}
\begin{figure}[!t]
	\centering
	\includegraphics[width=0.8\columnwidth]{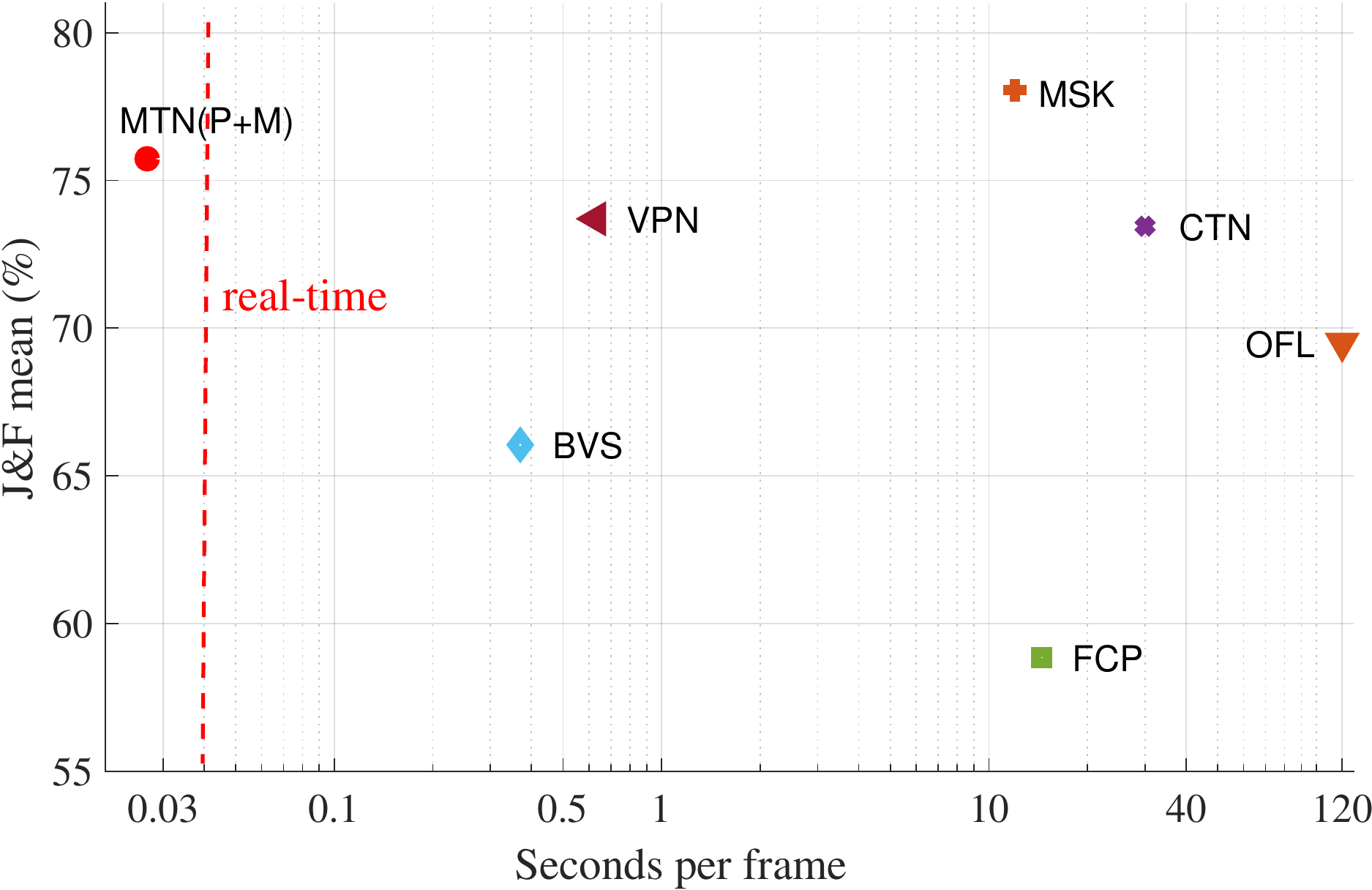}
	\caption{$\mathcal{J}\&\mathcal{F}$ mean versus runtime on DAVIS-16 trainval set.}
	\label{fig_davis2016_all}
	\vspace{-2ex}
\end{figure}

Notice that our model is only trained on  image object segmentation dataset (PASCAL VOC and MSRA10K) without any annotated video sequences. In order to demonstrate the robustness of our approach, we directly evaluate the proposed method MTN on the entire DAVIS-16 trainval set (\ie, 50 videos). We compare with all the available methods of high performance on this benchmark dataset\footnote{https://davischallenge.org/davis2016/soa\_compare.html, trainval set of the DAVIS-16 benchmark dataset.}, including MSK~\cite{CVPR2017_Perazzi}, CTN~\cite{CVPR2017_Jang}, VPN~\cite{CVPR2017_Jampani}, OFL~\cite{CVPR2016_Tsai}, BVS~\cite{CVPR2016_Nicolas}, FCP~\cite{ICCV2015_Perazzi} and JMP~\cite{SIGGRAPH2015_Fan}. MSK requires the first frame fine-tuning on each video sequence to achieve high accuracy. CTN, VPN, OFL, FCP and JMP require the optical flow to propagate the previous mask to the current frame. BVS segments the target object in a bilateral space with graph cut. 
The standard evaluation metrics~\cite{CVPR2016_Perazzi} include average region similarity $\mathcal{J}$, contour accuracy $\mathcal{F}$ and processing time. The results of those competitors are obtained from the corresponding published papers. 

As the processing time reported in \tab \ref{tbl_davis2016_all}, the proposed approach MTN only takes 0.027 seconds per frame for the single object segmentation, which is significantly faster than existing methods. In particular, MTN is about $440\times$ faster than MSK and $13 \times$ faster than BVS. Besides, \fig \ref{fig_davis2016_all} reports the overall performance on $\mathcal{J} \& \mathcal{F}$ mean and runtime, \ie, $(\mathcal{J}+\mathcal{F})/2$ and processing speed. It can be seen that only the proposed method can achieve a real-time speed. Besides, our method achieves the second best performance, which is also very close to MSK by a small margin of 2.35\% on $\mathcal{J} \& \mathcal{F}$ mean. 

As shown in \tab \ref{tbl_davis2016_all}, by training on the image segmentation dataset PASCAL VOC, our method MTN (P) can achieve an accuracy of 72.4\% on $\mathcal{J}$ and 74.9\% on $\mathcal{F}$. Besides, by increasing the diversity of training samples from MSRA10K dataset, the accuracy of our method MTN (P+M) is further improved by 2.9\% on $\mathcal{J}$ and 1.2\% on $\mathcal{F}$. Notice that even without the first frame fine-tuning and temporal cues for mask propagation, our method still achieves competitive performance when compared to the state-of-the-art methods. In particular, the $\mathcal{J}$ mean of MTN (P+M) is 75.3\%,  which is only less than the best method MSK which requires 12 seconds (\vs 0.027s of our method) to process one frame, and outperforms all the other methods. Moreover, in terms of contour accuracy $\mathcal{F}$, the proposed method MTN (P+M) achieves the best performance.

\begin{table}[]
	\centering
	\caption{Results ($\mathcal{J}$ mean metric) on the DAVIS-16 dataset with categories excluded from the PASCAL VOC dataset.}
	\resizebox{1\columnwidth}{!}{
	\begin{tabular}{l|c c c c c c c c}
		\hline
		& bear & bswan & camel & eleph & goat & malw & rhino & Avg. \\ \hline
		TFN~\cite{CVPR2016_Hong}  & 73.7 & 83.4  & 65.5  & 76.1  & 78.1 & 17.9 & 42.4  & 62.4 \\ \hline
		UOS~\cite{ACCV2018_Chen} &  89.8 & 76.7  & 72.0  & 73.8 & 83.3 & 41.6 & 71.0  & 72.6 \\ \hline
		MTN (P) &  {\bf 91.6} & {\bf 90.6}  & {\bf 80.3}  & {\bf 84.9}  & {\bf 83.9} & {\bf 89.2} & {\bf 91.6}  & {\bf 87.5} \\ \hline
	\end{tabular}
	}
	\label{tbl_unseen}
\end{table}
\subsection{Results on Unseen Objects in DAVIS-16 Dataset}
\label{subsec_unseen}

\begin{figure*}[!t]
	\centering
	\includegraphics[width=0.90\textwidth]{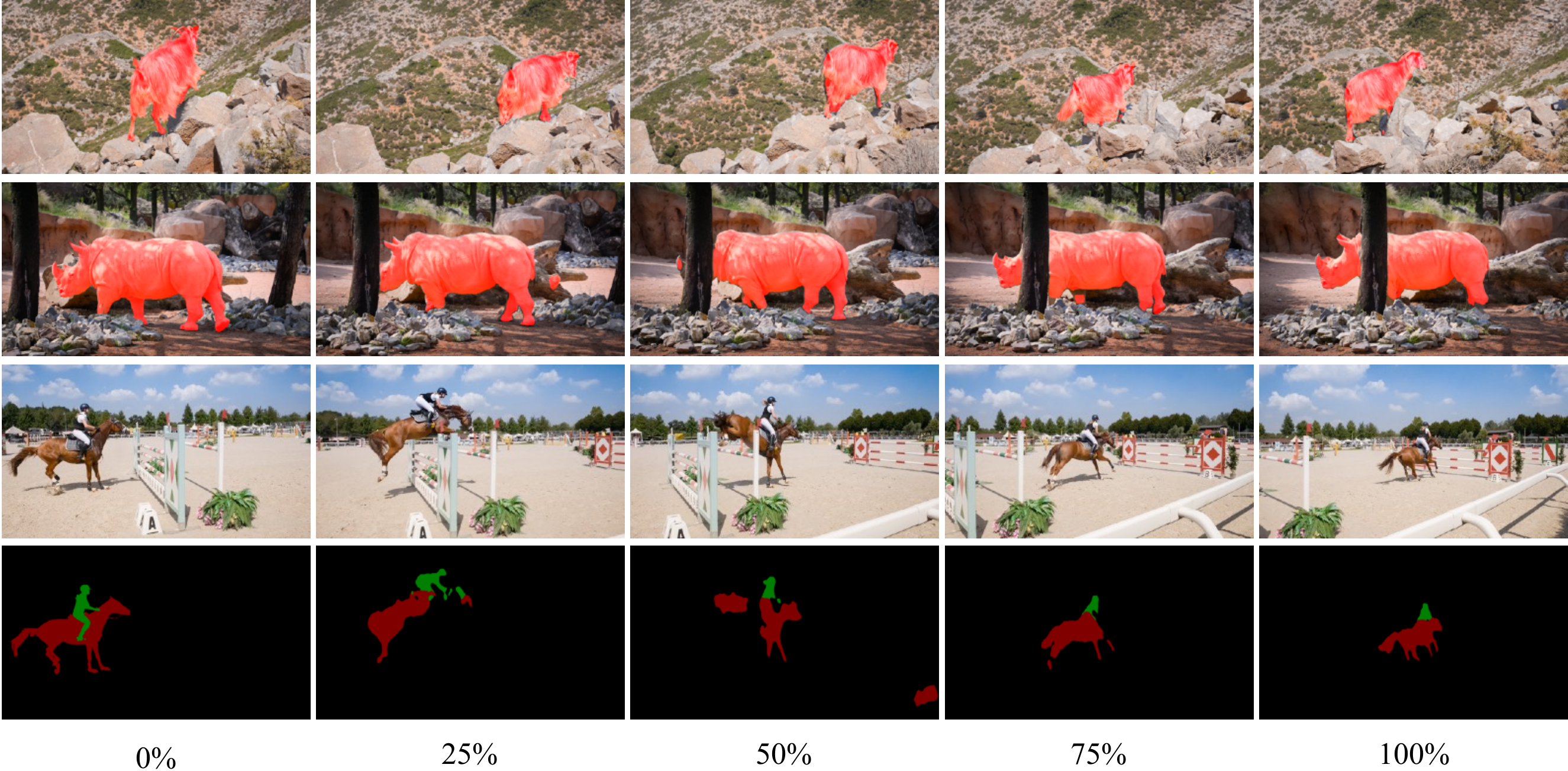}
	\caption{Example results of the proposed method. The first row and second row show the performance on unseen object category \emph{goat} and \emph{rhino}, respectively. The last two rows are results on the video \emph{horsejump-high} for multi-object segmentation.}
	\label{fig_results}
	\vspace{-3ex}
\end{figure*}

Because our approach does not rely on the particular object categories, it is robust to unseen object segmentation. Similar to the previous work UOS (unseen object segmentation in videos)~\cite{ACCV2018_Chen}, we also manually exclude all the videos with object categories that exist in the 20 categories of the PASCAL VOC dataset~\cite{IJCV2010_Everingham}. To evaluate our method, we compare with two mask transfer based approaches UOS~\cite{ACCV2018_Chen} and TFN~\cite{CVPR2016_Hong}, which focus on transferring the information of seen objects to unseen objects. As shown in \tab \ref{tbl_unseen}, our method achieves the best performance across all 7 videos. Two examples on unseen object categories \emph{goat} and \emph{rhino} are demonstrated in the first row and second row of \fig \ref{fig_results}, respectively. Furthermore, our method achieves a high accuracy of 87.5\% on $\mathcal{J}$ mean, which significantly improves the overall performance by 14.9\% on average. 

\begin{table}[!t]
	\centering
	\small
	\caption{Comparisons on the DAVIS-16 validation set (20 videos). FT, PF and OF are the same as in \tab \ref{tbl_davis2016_all}. }	
	\resizebox{1\columnwidth}{!}{
		\begin{tabular}{l|ccc|c|c|c}   \hline
			& FT   & PF      & OF   & Time (s) & $\mathcal{J}$ mean &  $\mathcal{F}$ mean    \\ \hline \hline
			OnAVOS \cite{BMVC2017_Voigtlaender} & Y   & Y       &   & 13     & 86.1 & 84.9   \\
%			MSK \cite{CVPR2017_Perazzi}         & Y   & Y       & Y    & 79.7 & 75.4 & 12       \\
			OSVOS \cite{CVPR2017_Caelles}       & Y   &         &   & 10     & 80.6 & 79.8       \\ 
			VM* \cite{ECCV2018_Hu}              &     &         &   & 0.17   & 79.2 & -      \\	
			\underline{\bf MTN}                 &     &         &   & \underline{0.027}   & \underline{75.9} & \underline{76.2}   \\
			BFVOS \cite{CVPR2018_Chen}          &     & Y       &   & 0.28   & 75.5 & 79.3   \\	
			OSMN \cite{CVPR2018_Yang}           &     & Y       &   & 0.14   & 74.0 & 72.9   \\  
%			RGMP   \cite{CVPR2018_Oh}           &     & Y       &      & 81.5 & 82.0 & 0.13     \\  
			RGMP*  \cite{CVPR2018_Oh}           &     &         &   & 0.13   & 73.5 & 74.2   \\
			CTN   \cite{CVPR2017_Jang}          &     & Y       & Y & 29.95  & 73.5 & 69.3   \\  
			OnAVOS* \cite{BMVC2017_Voigtlaender}&     & Y       &   & 3.55   & 72.7 & -      \\  
			VPN \cite{CVPR2017_Jampani}         &     & Y       & Y & 0.63   & 70.2 & 65.5   \\  
			PLM \cite{ICCV2017_Yoon}            & Y   & Y       &   & 0.30   & 70.0 & 62.5   \\ 
			BVS \cite{CVPR2016_Nicolas}         &     & Y       &   & 0.37   & 60.0 & 58.8   \\
			OSVOS* \cite{CVPR2017_Caelles}      &     &         &   & 0.10   & 52.5 & -      \\  \hline 
		\end{tabular}
		}
		\begin{tablenotes}  
			\footnotesize
			\item OnAVOS* and OSVOS*: without the first frame fine-tuning; VM* and RGMP*: without the previous frame for mask propagation.
		\end{tablenotes}  
	\label{tbl_davis2016}
\end{table}

\subsection{Results on the DAVIS-16 Validation Dataset}
\label{subsec_val}
In order to compare the proposed approach with more available state-of-the-art methods, we randomly choose a pair of video frames from the DAVIS-16 training set, and then fine-tune the pre-trained model (on the image segmentation PASCAL VOC and MSRA10K). Specifically, on the DAVIS-16 val set, we compare with a set of representative methods OnAVOS \cite{BMVC2017_Voigtlaender}, OSVOS \cite{CVPR2017_Caelles}, PLM \cite{ICCV2017_Yoon}, CTN \cite{CVPR2017_Jang}, BVS \cite{CVPR2016_Nicolas}, VPN \cite{CVPR2017_Jampani}, and the most recent approaches BFVOS \cite{CVPR2018_Chen}, OSMN \cite{CVPR2018_Yang}, RGMP \cite{CVPR2018_Oh} and VM \cite{ECCV2018_Hu}. Besides, to demonstrate the performance of compared algorithms with \emph{detection-based} strategy, we use RGMP \cite{CVPR2018_Oh} and VM \cite{ECCV2018_Hu} without the mask propagation component for fair comparison. 
In the following, we report both of the region similarity $\mathcal{J}$, contour accurayc $\mathcal{F}$ and processing time (obtained from the corresponding published papers) of the compared methods. 

\begin{table}[!t]
	\centering
	\small
	\label{tbl_davis2017}
	\caption{Comparisons on the DAVIS-17 validation set (30 videos). FT, PF and OF are the same as in \tab \ref{tbl_davis2016_all}.} 
%		* indicates a variation implementation of the corresponding algorithm. '-' denotes the corresponding values are not reported.}
	\resizebox{1\columnwidth}{!}{
		\begin{tabular}{l|ccc|c|c|c}  \hline
			& FT  & PF      & OF   & Time (s) & $\mathcal{J}$ mean & $\mathcal{F}$ mean \\ \hline \hline
			OnAVOS \cite{BMVC2017_Voigtlaender}  & Y   & Y    &        & 26   &  64.5   &   71.1       \\
%			VM   \cite{ECCV2018_Hu}              &     & Y       &      & 56.5  & - & 0.35    \\ \hline
			OSVOS  \cite{CVPR2017_Caelles}       & Y    &         &    & 20   & 52.1   & 62.1       \\		
			OSMN  \cite{CVPR2018_Yang}           &     & Y       &     & 0.28  & 52.5   & 57.1       \\
			MSK \cite{CVPR2017_Perazzi}          & Y   & Y       & Y   & 18    & 51.2  & 57.3       \\  
			\underline{\bf MTN}                  &     &         &     & \underline{0.048} & \underline{49.4} & \underline{59.0}   \\
			MSK* \cite{CVPR2017_Perazzi}         &     & Y       & Y   & 18 & 44.6  & 47.6    \\     
			OnAVOS* \cite{BMVC2017_Voigtlaender} &     & Y       &     & 7.10  & 39.5  & -    \\   
			OSVOS*  \cite{CVPR2017_Caelles}      &     &         &     & 0.20  & 36.4  & 39.5    \\
			 \hline  
		\end{tabular}
	}
	
	\begin{tablenotes}  
		\footnotesize
		\item OnAVOS*, OSVOS*, MSK*: without the first frame fine-tuning.
	\end{tablenotes}  
			
\end{table}
\begin{figure}[!t]
	\centering
	\includegraphics[width=0.8\columnwidth]{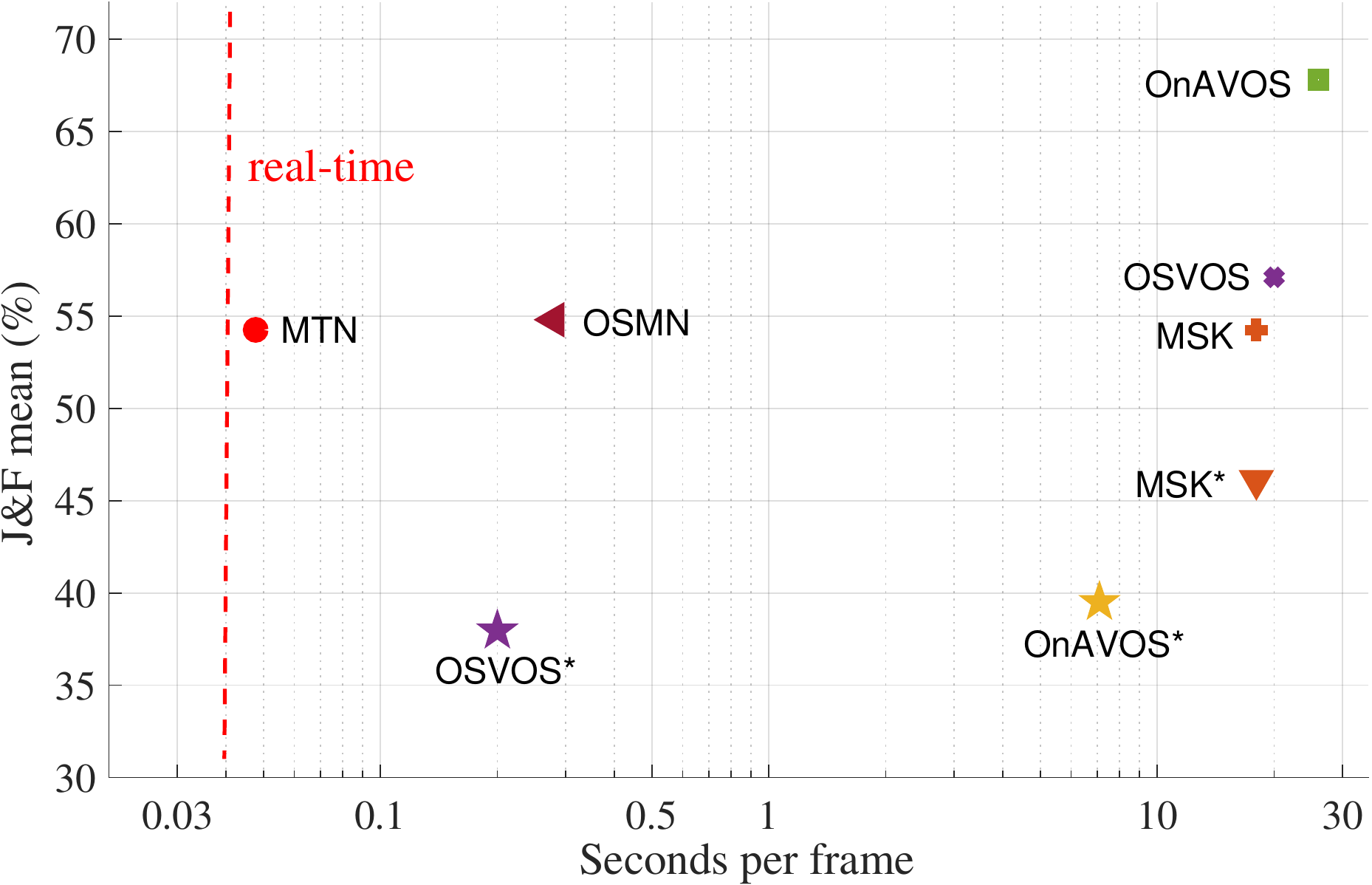}
	\caption{$\mathcal{J}\&\mathcal{F}$ mean versus runtime on DAVIS-17 validation set.}
	\vspace{-2ex}
	\label{fig_davis2017}
\end{figure}
As shown in \tab \ref{tbl_davis2016}, among all the methods which have not used the first frame fine tuning step, including VM*, BFVOS, OSMN, RGMP*, CTN, OnAVOS*, VPN, BVS and OSVOS*, the $\mathcal{J}$ mean of our method MTN is comparable, which is only not as good as VM* by a small margin, \ie 75.9\% vs 79.2\%. However, in terms of processing speed, MTN is about $6 \times$ faster than the matching-based method VM*, \ie, 0.027 vs 0.17. In addition, when compared to another matching-based algorithm BFVOS, our method is about $10\times$ faster and achieves a better $\mathcal{J}$ mean. It is worth mentioning that the accuracy of OSVOS and OnAVOS degrades a lot without the first frame fine-tuning, \ie, 79.8\% vs 52.5\% and 86.1\% vs 72.7\%, respectively.

\subsection{Results on Multi-object Segmentation}
\label{subsec_multi}
To validate the effectiveness of our method on multi-object segmentation, we carry out experiments on the DAVIS-17 val set, which consists of 30 videos with 61 different objects, and thus the average number of objects is 2 per video sequence. Since this dataset contains many similar objects in a video sequence, it is very challenging to obtain accurate segmentation results on this dataset. Therefore, the overall performance of all methods on this dataset is not as good as the single object segmentation, as shown in \tab \ref{tbl_davis2017}. \fig \ref{fig_davis2017} shows the overall performance on $\mathcal{J} \& \mathcal{F}$ mean and runtime. From the perspective of speed analysis, MTN only takes 0.048 seconds per frame during testing, \ie 20 fps. It is significantly faster than all the compared methods and close to a real-time speed of 25 fps. Besides, even without the first frame fine-tuning and mask propagation from previous frame, the performance of our method is still comparable to the state-of-the-art methods, \eg MTN outperforms the most recent method OSMN on $\mathcal{F}$ mean, \ie, 59.0\% vs 57.1\%. Without the first frame fine-tuning, the performance of OnAVOS*, OSVOS* and MSK* degrades a lot. For example, the $\mathcal{J}$ mean values of OnAVOS degrades from 64.5\% to 39.5\%. 

%Since many methods, such as MSK \cite{CVPR2017_Perazzi} and RGMP \cite{CVPR2018_Oh}, often concatenate the object mask with RGB image as a 4-channel input, when segmenting more objects, they need more processing time on the image and mask features extraction. For example, if there are $M$ target objects for segmentation, they need to extract $M$ times of both the image features and mask features, which is cumbersome and time-consuming. In contrast, our method extract the image features only once and mask feature for each target object. Therefore, our method is more efficient for multiple object segmentation. For example, the processing time is 0.08s rather than 0.10s for two object segmentation (reminder that the processing time is 0.05s for a single object).

\begin{table}[]
	\centering
	\caption{Runtime the of the main modules in MTN.}
	\resizebox{1\columnwidth}{!}{
    \begin{tabular}{c|c|c|c|c} \hline
    Module & \begin{tabular}[c]{@{}c@{}}Image\\  encoder\end{tabular} & \begin{tabular}[c]{@{}c@{}}Mask\\ encoder\end{tabular} & \begin{tabular}[c]{@{}c@{}}Pixel \\ matching\end{tabular} &  \begin{tabular}[c]{@{}c@{}}Bottom-up\\ decoder\end{tabular} \\  \hline
    Time (ms)  & 10    & 0.78       & 7.6    &  5.0  \\ \hline                 
    \end{tabular}
    }
    \label{tbl_time}
    \vspace{-2ex}
\end{table}

\subsection{Runtime Analysis} 
The proposed method is implemented and evaluated on a single NVIDIA GeForce 1080 Ti GPU, and it achieves 37 fps on images of $854 \times 480$ pixels. For detailed runtime analysis, we report the runtime of each module. As shown in \tab \ref{tbl_time}, we can see that the most time-consuming component is the image encoder module for image feature extraction. The processing time of other components has been greatly improved due to the efficient network design of MTN. To have  an intuitive impression, compared with the RGMP~\cite{CVPR2018_Oh} which runs on the same platform, our method is 4.8$\times$ faster than RGMP (less than 8 fps).  

\subsection{Qualitative Analysis}
To illustrate the effectiveness of our method, we present some representative examples for both of the unseen object segmentation and multi-object segmentation, as shown in Fig. \ref{fig_results}. The appearance of the object \emph{goat} in the first row is very similar to the background, and the second object \emph{rhino} is heavily occluded by the tree. It demonstrates that our method is robust to background clutter and heavy occlusion and can achieve good performance on unseen object segmentation. The last two rows show that our method can also be used for multi-object segmentation. 

\section{Conclusion}

In this paper, we report a novel mask transformation network (MTN) video object segmentation method, which can achieve a real-time processing speed with reasonable accuracy. MTN is developed based on the idea of mask transfer from the annotated object mask to the target frame. A global pixel matching method with a mask encoder network is proposed to accelerate the processing speed and ensure the segmentation accuracy. Experiments on the DAVIS datasets demonstrate that our method achieves a processing speed of 37 fps, which exceeds the real-time requirement of 25 fps. By training MTN on the image datasets only, it can achieve an accuracy of 75.3\% on the entire DAVIS-16 trainval set, which is comparable to the state-of-the-art methods. Besides, the proposed method does not rely on the particular object categories, significantly improving the overall performance on unseen object segmentation by 14.9\% on average. Moreover, in experiments, we also demonstrate that our method can be also used for multi-object segmentation.

\clearpage

{\small
\bibliographystyle{ieee}
\bibliography{egbib}
}

\end{document}